\documentclass[letterpaper, 10 pt, conference]{ieeeconf}

\IEEEoverridecommandlockouts
\usepackage[utf8]{inputenc}
\usepackage[T1]{fontenc}
\usepackage{amsmath}
\usepackage{amssymb}
\usepackage{graphicx}
\title{\LARGE \bf
Distributed Multi Robot Lunar Cargo Transportation via Phase Decomposed Reinforcement Learning
}

\author{Ashutosh Mishra$^{*1,2}$, Elian Neppel$^{1}$, Shreya Santra$^{1}$, Antoine Jonqui\`eres$^{3}$,\\ Muhammad Athallah Naufal$^{4}$,  Kentaro Uno$^{1}$, and Kazuya Yoshida$^{1}$%
\thanks{This work was supported by JST Moonshot R\&D Program, Grant Number JPMJMS223B.}%
\thanks{$^{1}$A. Mishra, E. Neppel, S. Santra, K. Uno, and K. Yoshida are with the Space Robotics Lab. (SRL), Department of Aerospace Engineering, Graduate School of Engineering, Tohoku University, Sendai 980--8579, Japan.  $^{2}$A. Mishra is also affiliated with the Dynamic Legged Systems (DLS) lab of the Istituto Italiano di Tecnologia (IIT), Genoa 16163, Italy. $^{3}$A. Jonqui\`eres is with \'Ecole Centrale de Lille, France. $^{4}$M. A. Naufal is with Institut Teknologi Bandung, Indonesia. $^{*}$Corresponding author: A. Mishra ({\tt\small ashutosh.mishra@dc.tohoku.ac.jp}).}%
}

\begin{document}
\bstctlcite{refs:BSTcontrol}

\maketitle
\thispagestyle{empty}
\pagestyle{empty}

\begin{abstract}
Modular reconfigurable robotic systems provide a scalable solution for cooperative surface operations in future lunar missions. However, cooperative cargo transportation remains challenging due to morphology-dependent topology changes, strong payload-induced coupling, long-horizon decision making, and safety constraints. This paper proposes a phase-decomposed reinforcement learning framework for cooperative cargo transport with distributed robotic units. The task is decomposed into lifting, transportation, and placement, each optimized with a dedicated joint-state policy capturing inter-agent coupling. Centralized training promotes stable convergence, while deployment uses onboard proprioception for control and OptiTrack motion capture for ground-truth evaluation and post-processed metrics. A deterministic phase controller expressed in Markov state representation regulates transitions between stages, and a failure-sensitive synchronization mechanism ensures coordinated progression and safety-aware halting during real-world execution. The framework is evaluated in simulation and through controlled field experiments at a JAXA space exploration test facility. Results demonstrate reliable cooperative transport across all stages in both simulation and hardware experiments.
\end{abstract}

\section{Introduction}
Sustained lunar surface missions require robotic systems capable of transporting construction materials, instruments, and logistical payloads under communication delay, terrain uncertainty, and strict safety constraints \cite{nasaartemis}. Modular reconfigurable robots are particularly suited to such environments, as they enable task-specific morphologies assembled from reusable units, supporting repairability, redundancy, and functional adaptability \cite{whitman_learning_2023, iwata_adaptive_2021}. 

Modularity introduces configuration-dependent kinematics, actuation topology, and inter-module coupling, reducing policy transferability across morphologies. In cooperative transport, additional coupling arises through the shared payload, producing strong dynamic interactions.

Existing decentralized reinforcement learning approaches allow distributed modules to act based on local observations while contributing to global objectives \cite{busoniu_decentralized_2006, wang_distributed_2022}. Centralized training with decentralized execution improves convergence stability in multi-agent settings \cite{sartoretti_distributed_2019, guo_decentralized_2023}. Deep reinforcement learning has shown effectiveness for cooperative navigation and transport \cite{zhang_decentralized_2020, han_reinforcement_2022}. However, most formulations assume fixed topology or optimize a single controller for structurally static teams, without addressing physical reconfiguration and topology-dependent coupling during task execution. For modular systems undergoing physical reconfiguration and multi-stage contact transitions, unified policy optimization can induce gradient interference across heterogeneous dynamics and fails to explicitly model configuration-dependent coupling \cite{yu_gradient_surgery_2020, whitman_learning_2023}.

\begin{figure}[t]
    \centering
    \includegraphics[width=\columnwidth]{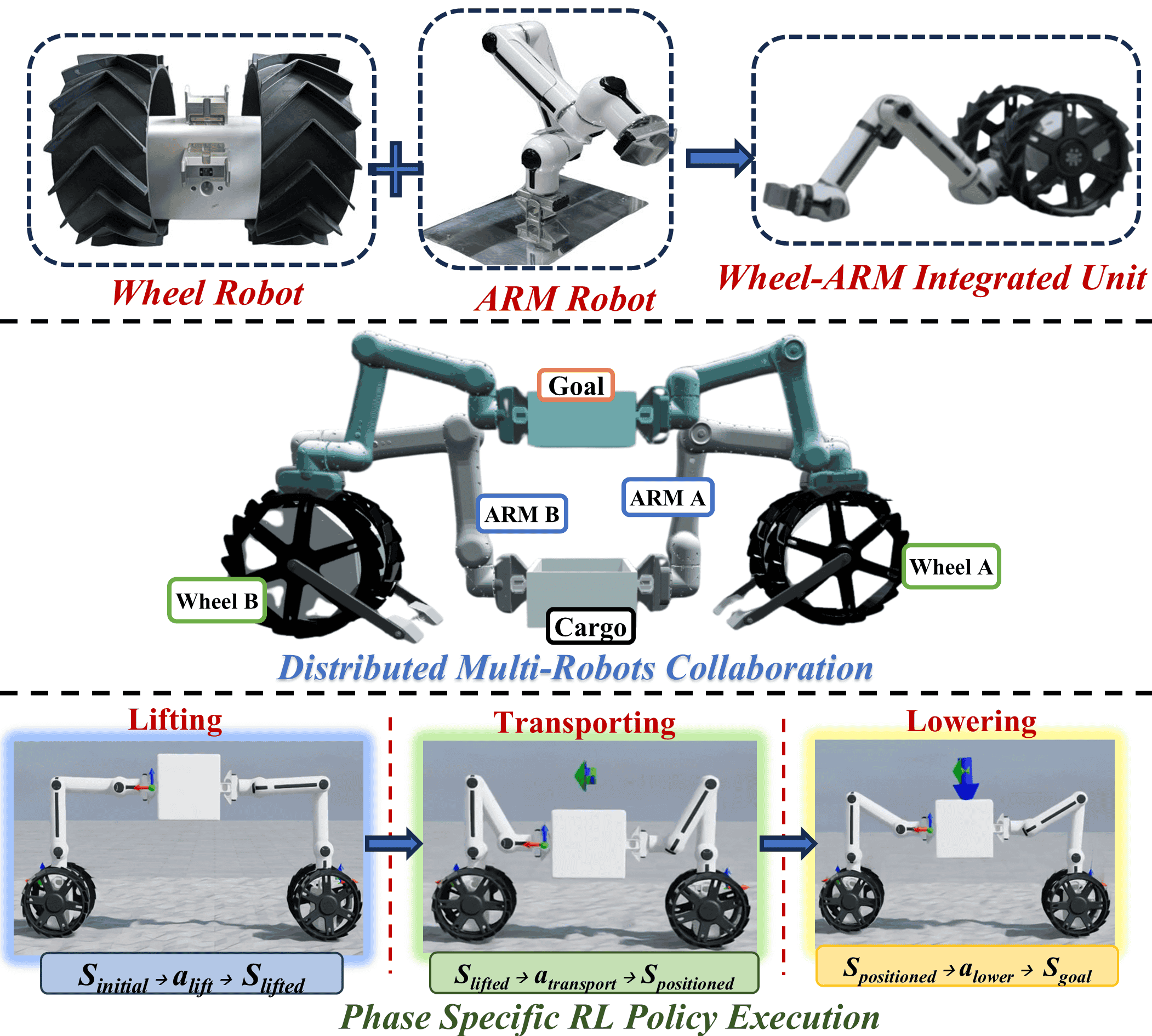}
    \caption{
    System overview of the proposed modular multi-phase reinforcement learning framework. Two distributed wheel arm integrated units cooperatively transport a shared payload through lifting, transportation, and placement stages.
    }
    \label{fig:system_overview}
\end{figure}

In cooperative lunar cargo transport, the problem is compounded by sequential operational stages. Lifting, transportation, and placement involve distinct contact conditions and objective structures. Furthermore, modules must coordinate differently depending on configuration, for example, synchronized actuation during lifting, compliant steering during transit, and precise alignment during placement.

To address this, we propose a phase-decomposed reinforcement learning framework for cooperative lunar cargo transport. The mission is decomposed into lifting, transportation, and placement stages, each optimized using a dedicated joint-state policy. Execution remains distributed across modules. In this study the modular robots are assembled into a transport-capable configuration and the framework evaluates coordinated operation under this configuration. A high-level Markov decision process governs stage transitions, while a synchronization layer coordinates inter-module execution and enforces safety-aware halting. The framework is validated on our in-house modular reconfigurable robotic platform~\cite{moonbot} through simulation and controlled experiments at a JAXA lunar-analog facility \cite{jaxafield}.

The key contributions are:

\begin{enumerate}
    \item Phase-structured reinforcement learning for cooperative transport under configuration-dependent dynamics and payload-induced coupling, separating policy optimization across distinct contact regimes.
    \item Centralized joint-state training with distributed execution for mechanically coupled yet physically independent modular robots.
    \item Discrete MDP-based phase gating with synchronization-aware safety coordination for sequential multi-stage execution.
    \item Simulation and lunar-analog hardware validation on a modular wheel–arm robotic system assembled for cooperative cargo transport.
\end{enumerate}

Beyond integrating established components, the central methodological contribution is the phase-structured formulation itself: a long-horizon, contact-rich cooperative-transport task is decomposed into phase-specific MDPs with dedicated reward geometries and coupled by a deterministic, synchronization-aware phase gate that keeps multi-stage sim-to-real execution stable and safety-bounded. This structure is not merely organizational. A monolithic policy trained over the same joint state and action space fails to converge (Sec.~\ref{subsec:baseline_no_decomp}), indicating that the decomposition mitigates gradient interference across heterogeneous contact regimes rather than re-packaging PPO, centralized training with decentralized execution, and MDP gating.

\section{Related Work}
Multi-robot collaboration for space and planetary applications has focused on modular and  distributed logistics, coordinated construction, and collective manipulation under communication constraints \cite{wang2025collaborative, polizzi2023multi}. Model-based decentralized transport strategies have demonstrated feasibility in structured environments \cite{farivarnejad_fully_2021, wang_model-based_2021}. However, these approaches depend on accurate dynamic models and pre-defined coordination logic.

Multi-agent reinforcement learning provides an alternative framework for adaptive coordination. Early decentralized learning formulations established theoretical foundations for distributed control \cite{busoniu_decentralized_2006}. Recent multi-agent reinforcement learning emphasizes centralized training with decentralized execution \cite{wang_distributed_2022, gronauer_multi-agent_2022}. Applications to decentralized navigation and cooperative transport demonstrate scalability benefits \cite{zhang_decentralized_2020, han_reinforcement_2022}. Recent work also investigates communication-efficient decentralized training mechanisms \cite{guo_decentralized_2023, nair_decentralizing_2022}. Despite these developments, most studies optimize a single policy across heterogeneous task phases.

Modular robotics introduces additional structural variability. Learning-based control generalization across reconfigurable morphologies has been explored to enhance adaptability \cite{whitman_learning_2023, iwata_adaptive_2021}. Morphology-aware co-optimization strategies have further improved transferability across configurations \cite{gao_co-optimizing_2025}. Robust synchronization strategies for modular systems have addressed distributed actuation coordination \cite{neppel2025robustmodularmultilimbsynchronization}. However, the structured decomposition of cooperative transport into phase-specific reinforcement learning problems, combined with synchronization-governed stage transitions, has not been systematically investigated.

The present work addresses this gap by introducing a phase-structured reinforcement learning architecture that integrates joint-state training, distributed execution, and safety-aware synchronization for cooperative lunar cargo transport.

\section{Modular Robotic System}
The robotic platform considered in this work~\cite{moonbot} consists of physically distinct locomotion and manipulation modules that can be mechanically combined to realize task specific configurations. The architecture separates mobility and manipulation at the hardware level, enabling structural reconfiguration according to operational requirements.

\subsection{Wheel Module}
The locomotion subsystem is implemented as a differential drive wheel module. The module comprises a rigid base with two independently actuated wheels, enabling planar motion through differential velocity control. The kinematic structure is nonholonomic and constrained to planar translation and rotation about the vertical axis. The module provides mobility for transport and positioning tasks and serves as the base platform when coupled with a manipulation module.

\subsection{Arm Module}
The manipulation subsystem is realized as a serial articulated arm with seven degrees of freedom. The kinematic chain enables three dimensional end effector positioning and orientation control. Each joint is independently actuated, allowing coordinated motion for grasping, lifting, and payload stabilization tasks. The arm module can operate independently or be mechanically attached to a mobility base.

\subsection{Wheel and Arm Integrated Configuration}
A mobile manipulation unit is formed by mechanically coupling one differential drive wheel module with one seven degree of freedom arm module. In this configuration, the wheel base provides planar mobility while the arm enables object interaction. The combined system exhibits coupled base and manipulator dynamics during motion and load handling. For cooperative cargo transportation, multiple integrated units physically engage a shared payload. The overall system architecture and phase-structured task execution are illustrated in Fig.~\ref{fig:system_overview}. The present study evaluates the proposed framework on a transport-capable assembly of the modular system; investigating policy transfer across alternative robot configurations is an important direction for future work. At the module level, learned policies can control new morphologies and generalize to previously unseen assemblies, as we demonstrate in prior work~\cite{mishra_isparo_2025}. The arms establish contact and generate interaction forces, while the wheel bases provide coordinated locomotion.

\section{Modular Multi-Phase Reinforcement Learning Policies}\label{sec:learning_policies}
The cooperative cargo transport problem is formulated as a phase-structured sequential decision-making process composed of three operational stages: lifting ($L$), transportation ($T$), and placement ($P$). The execution is sequential, whereas policy learning remains modular. Each stage is optimized independently over a shared joint state–action interface, enabling replacement or refinement of individual policies without retraining the full framework.

\subsection*{System-Level Formulation}
Let $N$ denote the number of integrated wheel–arm units. The coupled system state at time $t$ is defined as

\begin{equation}
s_t =
\left[
x_t^{(1)}, \dots, x_t^{(N)}, x_t^{c}
\right],
\end{equation}

where $x_t^{(i)}$ represents the pose, velocity, and joint configuration of robot $i$, and $x_t^{c}$ denotes the pose and velocity of the cargo.

The joint action vector is

\begin{equation}
a_t =
\left[
u_t^{(1)}, \dots, u_t^{(N)}
\right],
\end{equation}

where $u_t^{(i)}\in\mathbb{R}^{9}$ contains the two differential-drive wheel velocity commands and the seven arm-joint position commands of robot $i$; for the $N{=}2$ transport-capable configuration the joint action is therefore $a_t\in\mathbb{R}^{18}$.

Each operational phase $k \in \{L, T, P\}$ is modeled as a Markov Decision Process

\begin{equation}
\mathcal{M}_k = (\mathcal{S}, \mathcal{A}, \mathcal{P}, r_k, \gamma),
\end{equation}

where $\mathcal{S}$ and $\mathcal{A}$ are joint state and action spaces, $\mathcal{P}$ denotes transition dynamics, $r_k$ is the phase-specific reward, and $\gamma \in (0,1)$ is the discount factor. Policies are trained under full joint-state observation and executed using local observations with shared cargo state.

\subsection*{Observation Model}

During centralized training:

\begin{equation}
o_t = s_t.
\end{equation}

During distributed execution, robot $i$ observes

\begin{equation}
o_t^{(i)} =
\left[
x_t^{(i)},
\Delta x_t^{c},
\Delta x_t^{(j)}
\right],
\end{equation}

where $\Delta x_t^{c}$ denotes relative cargo pose and $\Delta x_t^{(j)}$ encodes relative neighboring robot states.

Because the system is assembled from a fixed set of known modules, the per-unit observation $o_t^{(i)}$ and action $u_t^{(i)}$ have fixed, predetermined dimensions, so the policy input and output dimensions are consistent between training and deployment.

\subsection*{Stage I: Lifting Policy}

The lifting stage aims to raise the cargo to a target height $h^\ast$ while maintaining balanced load distribution and minimizing oscillatory behavior.

The reward function is
\begin{equation}
\begin{aligned}
r_L =\;&
- w_{h}\Big(\frac{h_t-h^\ast}{h_{\text{tol}}}\Big)^2
- w_{\dot h}\Big(\frac{\dot h_t}{\dot h_{\text{tol}}}\Big)^2 \\
&- w_{f}\Big(\frac{\sigma_f}{\sigma_{f,\text{tol}}}\Big)
- w_{\theta}\Big(\frac{\theta_{\text{tilt}}}{\theta_{\text{tol}}}\Big)^2 .
\end{aligned}
\end{equation}

where $h_t$ denotes cargo height, $\dot{h}_t$ vertical velocity, 
$\sigma_f$ the variance of interaction forces across arms, 
$\theta_{\text{tilt}}$ the cargo tilt angle, and 
$w_h, w_{\dot h}, w_f, w_\theta > 0$ are weighting coefficients. The inclusion of $\sigma_f$ explicitly enforces balanced cooperative lifting across wheel–arm units.
\noindent\textbf{Force signal used for $\sigma_f$.}
In simulation, interaction forces at each arm-payload contact were obtained directly from the physics engine contact wrench and $\sigma_f$ was computed across arms at each step.
In hardware, direct contact force sensing was not available; therefore, $\sigma_f$ was used only during simulation training as a regularizer to promote balanced load sharing, while deployment relied on encoder-based state feedback and the synchronization clamp to suppress asymmetric transients.

\subsection*{Stage II: Transportation Policy}

The transportation stage governs cooperative planar motion from $p_A$ to $p_B$ while preserving payload stability.

The reward is defined as

\begin{equation}
\label{eq:reward_transport}
\begin{aligned}
r_T =\;&
- w_p \Big(\frac{\|p_t - p_B\|}{p_{\mathrm{tol}}}\Big)^2
- w_v \Big(\frac{\|\dot{p}_t - v_{\mathrm{cmd}}\|}{v_{\mathrm{tol}}}\Big)^2 \\
&- w_\theta \Big(\frac{\theta_{\mathrm{tilt}}}{\theta_{\mathrm{tol}}}\Big)^2
- w_h \Big(\frac{h_t - h^\ast}{h_{\mathrm{tol}}}\Big)^2 .
\end{aligned}
\end{equation}

where $p_t$ denotes cargo planar position, $\dot{p}_t$ planar velocity, 
$v_{\text{cmd}}$ the commanded transport velocity, 
$\theta_{\text{tilt}}$ enforces upright stability, and 
$h_t$ preserves lifting height. 
This formulation decouples translational tracking from stability and height regulation.

\subsection*{Stage III: Placement Policy}

The placement stage ensures controlled lowering and precise alignment at the target configuration.

The reward is
\begin{equation}
\begin{aligned}
r_P =\;&
- w_{p}\Big(\frac{\|p_t-p_B\|}{p_{\text{tol}}}\Big)^2
- w_{h0}\Big(\frac{h_t}{h_{0,\text{tol}}}\Big)^2 \\
&- w_{\dot h}\Big(\frac{\dot h_t}{\dot h_{\text{tol}}}\Big)^2
- w_{\theta}\Big(\frac{\theta_{\text{tilt}}}{\theta_{\text{tol}}}\Big)^2 .
\end{aligned}
\end{equation}

where $h_t$ penalizes residual height, $\dot{h}_t$ ensures smooth descent, 
$\theta_{\text{tilt}}$ enforces upright alignment, and $w_p, w_{h0}, w_{\dot h}, w_\theta > 0$ are weighting coefficients.

\subsection*{Policy Optimization}

Each phase policy $\pi_{\theta_k}$ maximizes expected discounted return

\begin{equation}
J(\theta_k)
=
\mathbb{E}_{\pi_{\theta_k}}
\left[
\sum_{t=0}^{T_k}
\gamma^t r_k(s_t, a_t)
\right],
\end{equation}

where $T_k$ denotes the phase horizon.

Policies are trained independently under centralized observation and deployed in distributed execution governed by a higher-level phase MDP and synchronization layer.

\section{Policy Coordination and Sim-to-Real Deployment}
\label{sec:sim2real}

Training in simulation yields stage-specific policies for lifting, transportation, and placement. However, direct deployment can be destabilized by actuator latency, wheel slip, contact uncertainty, and state-estimation noise, due to distributed modular hardware with dynamically coupled subsystems. To reduce this sim-to-real discrepancy, execution is governed by a hybrid coordination layer composed of (i) a discrete phase MDP that gates the active policy and (ii) a joint synchronization mechanism that bounds per-step progression of the coupled wheel--arm--cargo--arm--wheel system.

\subsection{Phase Coordination via MDP Gating}

We augment the continuous system state $s_t$ with a discrete phase variable
\begin{equation}
m_t \in \mathcal{M}\triangleq\{0,1,2,3\}, \qquad x_t \triangleq (s_t,m_t),
\label{eq:hybrid_mode_state}
\end{equation}
where $0$ denotes Idle, $1$ Lift, $2$ Transport, and $3$ Place.

\begin{figure}[t]
  \centering
  \includegraphics[width=\columnwidth]{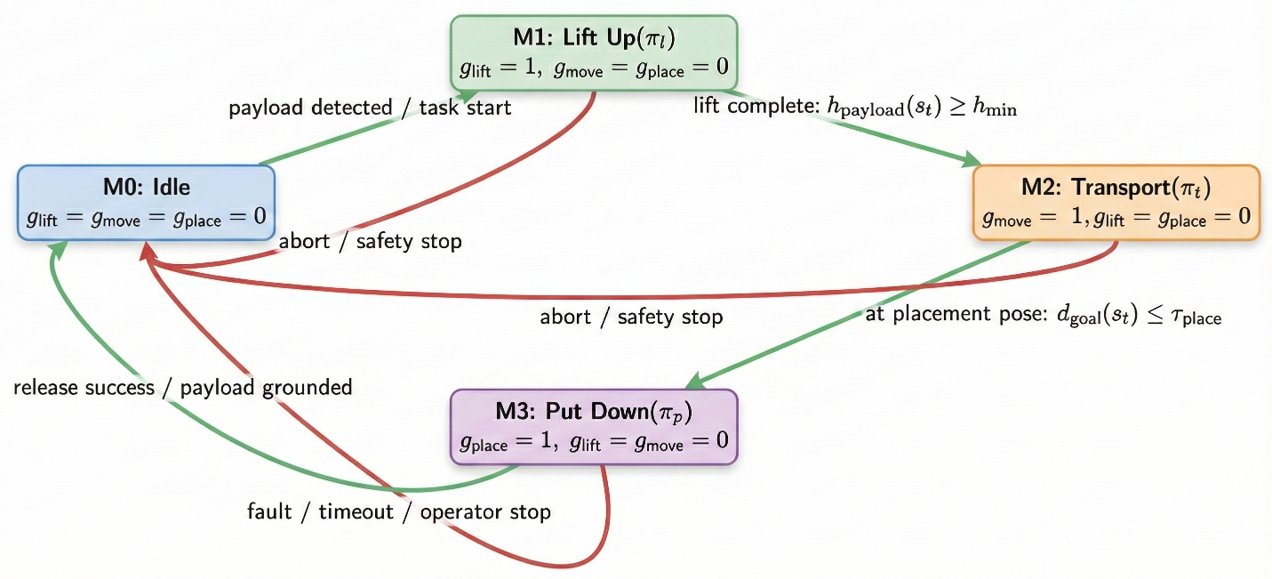}
  \caption{Phase MDP gating for multi-stage cooperative cargo transport. The discrete mode variable $m_t \in \{0,1,2,3\}$ activates exactly one phase policy at a time through indicator gating, with $M0$ denoting Idle, $M1$ Lift ($\pi_\ell$), $M2$ Transport ($\pi_t$), and $M3$ Put Down ($\pi_p$). The gating variables satisfy $(g_{\text{lift}},g_{\text{move}},g_{\text{place}})\in\{(1,0,0),(0,1,0),(0,0,1)\}$ for active phases and $(0,0,0)$ in Idle. Green transitions denote task-driven progression, whereas red transitions indicate abort or safety-stop overrides that return the system to Idle.}
  \label{fig:cargo_mdp}
\end{figure}

As illustrated in Fig.~\ref{fig:cargo_mdp}, the discrete mode variable gates the active stage policy through $(g_{\text{lift}}, g_{\text{move}}, g_{\text{place}})$, enabling sequential progression with safety-stop overrides.

Only one policy is active at any time, selected by indicator gating:
\begin{equation}
a_t
=
\mathbb{I}[m_t=1]\pi_{\ell}(s_t)
+
\mathbb{I}[m_t=2]\pi_{t}(s_t)
+
\mathbb{I}[m_t=3]\pi_{p}(s_t).
\label{eq:gated_action_compact}
\end{equation}

Phase transitions follow measurable task events with a safety override:
\begin{equation}
m_{t+1}=
\begin{cases}
0, & \sigma_t=1,\\
1, & m_t=0 \wedge \mathrm{start}(s_t)=1,\\
2, & m_t=1 \wedge h_{\mathrm{payload}}(s_t)\ge h_{\min},\\
3, & m_t=2 \wedge d_{\mathrm{goal}}(s_t)\le \tau_{\mathrm{place}},\\
0, & m_t=3 \wedge \mathrm{released}(s_t)=1,\\
m_t, & \text{otherwise}.
\end{cases}
\label{eq:mode_transition}
\end{equation}
Here $\sigma_t$ denotes a safety-stop signal (fault, timeout, or operator intervention), enforcing immediate halt and reset to Idle.

\subsection{Joint Synchronization for Cargo Mode}

When the two units carry a shared payload, they form a closed kinematic chain,
\[
\texttt{Wheel}_1 - \texttt{Arm}_1 - \texttt{Cargo} - \texttt{Arm}_2 - \texttt{Wheel}_2 ,
\]
so uncoordinated motion of any one subsystem is resisted by the others through the rigid cargo. This layer therefore admits as much of each policy command as possible while ensuring that no subsystem advances far ahead of the others, by clamping the commanded motion to a small admissible region around the current sensed state at every control step $k$.

We stack the sensed states of the two arms and two wheel bases at step $k$ into a single vector
\begin{equation}
Y_k \triangleq
\begin{bmatrix}
g_k^{(A_1)}\\
g_k^{(A_2)}\\
p_k^{(W_1)}\\
p_k^{(W_2)}
\end{bmatrix},
\qquad
g_k^{(A_i)}\in SE(3),\; p_k^{(W_j)}\in\mathbb{R}^2,
\label{eq:stacked_state}
\end{equation}
where $g_k^{(A_i)}$ is the pose of arm $i$ and $p_k^{(W_j)}$ the planar position of wheel base $j$. The policy output $a_k$ defines a desired increment that, applied to the previous command $U_{k-1}$, produces the raw target state
\begin{equation}
F_k \triangleq U_{k-1}\oplus \Delta(U_{k-1},a_k),
\label{eq:target_state}
\end{equation}
where $\oplus$ denotes composition on the $SE(3)$ (arm) components and addition on the planar wheel components.

Rather than issuing $F_k$ directly, we require the commanded stacked state to remain within an admissible set centered on the sensed state $Y_k$:
\begin{equation}
\mathcal{V}_k \triangleq 
\{ X \mid d(X,Y_k) \le 1 \},
\label{eq:admissible_set}
\end{equation}

\begin{equation}
\begin{aligned}
d(X,Y) \triangleq 
\max \Big(
& d_{SE}(X^{(A_1)},Y^{(A_1)}), \\
& d_{SE}(X^{(A_2)},Y^{(A_2)}), \\
& \| (X^{(W_1)}-Y^{(W_1)}) \oslash \Delta_w \|_2, \\
& \| (X^{(W_2)}-Y^{(W_2)}) \oslash \Delta_w \|_2
\Big).
\end{aligned}
\label{eq:stacked_metric_clean}
\end{equation}
where the distance $d$ is the worst case (an $\ell_\infty$ maximum) over the four subsystems, each normalized by its own tolerance, and $\oslash$ is elementwise division by the per-axis tolerance $\Delta_w$, so that a wheel term reaches $1$ when the combined normalized planar deviation attains unit magnitude. The arm term uses a normalized pose distance
\begin{equation}
d_{SE}(g_1,g_2)=\max\Big(\frac{\|t_1-t_2\|_2}{d_{\max}},\frac{\angle(R_1^\top R_2)}{\theta_{\max}}\Big),
\end{equation}
with $g=(R,t)\in SE(3)$ and $\angle(\cdot)$ the rotation angle, which reaches $1$ when either the translation error hits $d_{\max}=0.03$ m or the rotation error hits $\theta_{\max}=5^\circ$. The wheel deviation bound is $\Delta_w=[0.02,\,0.02]^\top$ m. Consequently $d(\cdot,Y_k)\le 1$ holds only if every subsystem stays within its tolerance simultaneously.

Finally, let $\mathcal{T}(t,S_k,F_k)$ interpolate from the previous command $S_k\triangleq U_{k-1}$ to the target $F_k$ as $t$ ranges over $[0,1]$. We issue the largest feasible step toward the target that still satisfies the admissibility constraint:
\begin{equation}
\begin{aligned}
t_k^{\max} &\triangleq \max \Big\{ t \in [0,1] \;\Big|\;
\mathcal{T}(t,S_k,F_k)\in \mathcal{V}_k \Big\}, \\
U_k &\triangleq \mathcal{T}\!\left(t_k^{\max}, S_k, F_k\right).
\end{aligned}
\end{equation}

Because the admissibility constraint is an $\ell_{\infty}$ maximum over subsystems, the subsystem closest to its per-step deviation bound caps $t_k^{\max}$ and hence the progress of the whole assembly, preventing the asynchronous arm or wheel motion that would otherwise induce internal cargo stress. If $t_k^{\max}$ is undefined (no feasible interpolation) or a safety stop is active ($\sigma_k=1$), the controller defaults to a hold command, thereby ensuring bounded deviation and safe stabilization.

\section{Experimental Setup}

\subsection{Simulation Environment and Training Configuration}

\begin{figure}[t]
    \centering
    \includegraphics[width=\columnwidth]{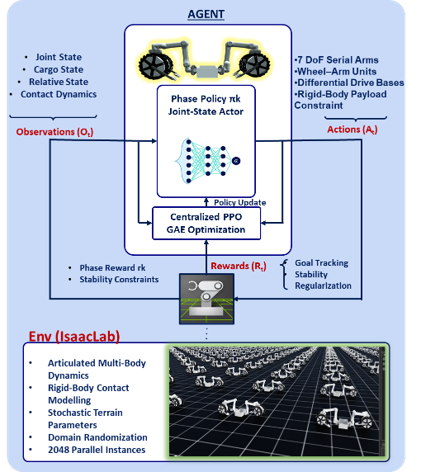}
    \caption{IsaacLab simulation and centralized PPO training architecture for phase-specific joint-state policies under articulated multi-body contact dynamics and domain randomization.}
    \label{fig:isaaclab_training}
\end{figure}

All phase-specific policies were trained in NVIDIA IsaacLab \cite{mittal2025isaaclab} using articulated rigid-body simulation with contact dynamics, as illustrated in Fig.~\ref{fig:isaaclab_training}. The cooperative system consisted of two integrated wheel–arm units mechanically coupled through a rigid payload in the configuration Wheel$_1$–Arm$_1$–Cargo–Arm$_2$–Wheel$_2$.

The robot model was defined via URDF and deployed in USD-based scenes within Isaac Sim. Each training scene contained a contact-enabled planar ground surface and illumination sources. To improve sample efficiency, 2048 to 4096 parallel environments were instantiated during training depending on the phase (Table~\ref{tab:ppo_params}).

The policy and value functions were implemented as separate multilayer perceptrons with ELU activations, optimized with PPO. The hidden-layer sizes are $[256, 128, 64]$ for the lifting and placement policies and $[128, 128, 128]$ for the transportation policy.

Moderate domain randomization was introduced, including bounded friction variation, terrain perturbations, actuation noise, and initial joint variation.
Domain randomization ranges were: wheel-ground friction coefficient $\mu\sim \mathcal{U}(0.6,1.0)$,
terrain heightfield perturbations with peak-to-peak amplitude $\leq 0.02$ m and spatial correlation length in $[0.05,0.15]$ m,
wheel velocity command noise $\epsilon_\omega\sim\mathcal{N}(0,0.05^2)$ rad/s,
and arm joint command noise $\epsilon_q\sim\mathcal{N}(0,0.01^2)$ (in joint units).

\begin{table}[t]
\centering
\caption{Per-phase PPO training configuration. The learning rate is the initial value under an adaptive schedule.}
\footnotesize
\renewcommand{\arraystretch}{1.1}
\setlength{\tabcolsep}{4pt}
\begin{tabular}{l c c c}
\hline
\textbf{Parameter} & \textbf{Lifting} & \textbf{Transport} & \textbf{Placement} \\
\hline
Algorithm & PPO & PPO & PPO \\
Learning rate & $1\times10^{-4}$ & $1\times10^{-3}$ & $1\times10^{-4}$ \\
Discount factor $\gamma$ & 0.98 & 0.99 & 0.98 \\
GAE parameter $\lambda$ & 0.95 & 0.95 & 0.95 \\
Clip ratio & 0.2 & 0.2 & 0.2 \\
Epochs per update & 5 & 5 & 5 \\
Mini-batches & 4 & 4 & 4 \\
Entropy coefficient & 0.006 & 0.005 & 0.006 \\
Parallel environments & 2048 & 4096 & 2048 \\
\hline
\end{tabular}
\label{tab:ppo_params}
\end{table}

\begin{table}[t]
\caption{Reward tolerances and weights used in all experiments.}
\label{tab:reward_weights}
\centering
\begin{tabular}{lcc}
\hline
Term & Tolerance & Weight \\
\hline
Height error $h_{\text{tol}}$ & $0.03$ m & $w_h=1.0$ \\
Vertical speed $\dot h_{\text{tol}}$ & $0.10$ m/s & $w_{\dot h}=0.5$ \\
Tilt $\theta_{\text{tol}}$ & $10^\circ$ ($0.174$ rad) & $w_\theta=0.7$ \\
Planar position $p_{\text{tol}}$ & $0.20$ m & $w_p=1.0$ \\
Planar speed $v_{\text{tol}}$ & $0.05$ m/s & $w_v=0.5$ \\
Force-variance $\sigma_{f,\text{tol}}$ & $1.0$ (normalized) & $w_f=0.3$ \\
Final height $h_{0,\text{tol}}$ & $0.02$ m & $w_{h0}=1.0$ \\
\hline
\end{tabular}
\end{table}
The per-phase training configuration is summarized in Table~\ref{tab:ppo_params}.

\subsection{Hardware Platform and Field Setup}

Hardware validation of the proposed multi-phase framework was conducted at JAXA’s Advanced Facility for Space Exploration in a controlled lunar-analog environment. Experiments were performed on a regolith-like sand test field designed to emulate lunar surface conditions. The terrain consisted of loose granular soil with non-uniform compaction and mild surface irregularities, introducing wheel slip, contact uncertainty, and uneven load distribution. This environment enables evaluation under terrain-induced disturbances.

The experimental platform comprised four independently operable robots from our modular reconfigurable system: two wheel bases and two articulated arms. Each wheel–arm pair was physically assembled into a mobile manipulation unit, and two such units engaged a shared payload. Although physically distributed and reconfigurable, the coupled system was trained under centralized joint-state observation and executed in a distributed manner, validating the framework under real hardware reconfiguration. Each unit combined a differential drive wheel base for planar locomotion with a seven degree of freedom articulated manipulator, instrumented with motor encoders for joint position and velocity measurement. The two manipulators engaged the payload from opposing sides, forming a mechanically coupled system throughout execution. No external structural support was provided.

To evaluate robustness across mass and geometry variations, three payload configurations were tested: a \textbf{Sledge} (5.3~kg elongated structure), a \textbf{Cargo box} (7.5~kg), and \textbf{Cargo with bricks} (9.5~kg).

\begin{figure}[t]
\centering
\includegraphics[width=\columnwidth]{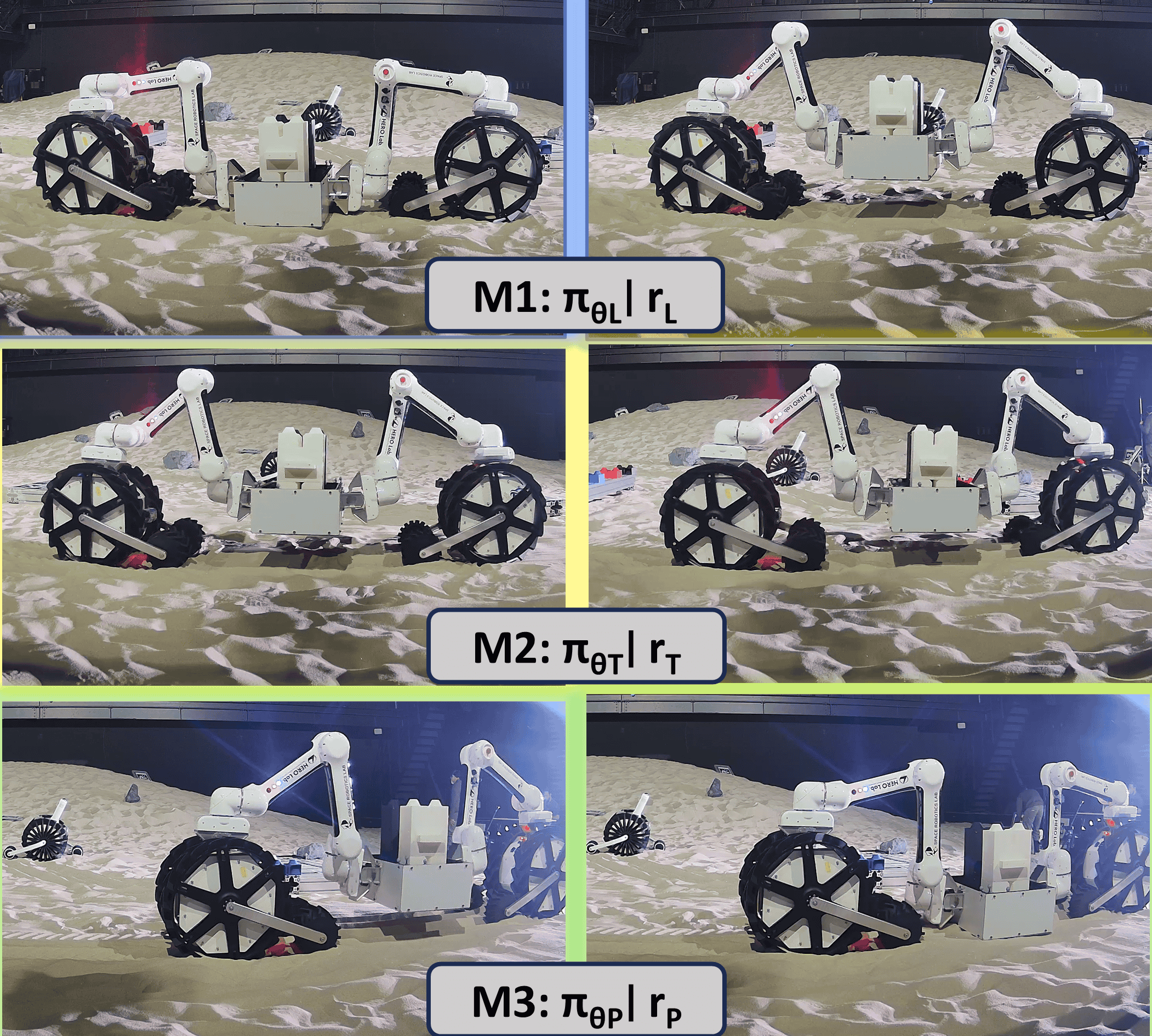}
\caption{Sequential hardware execution of lifting (M1), transportation (M2), and placement (M3) during cooperative cargo transport in a lunar-analog field experiment. Each row corresponds to a phase-specific policy $\pi_{\theta_k}$ with reward $r_k$, activated through discrete MDP gating under distributed execution.}
\label{fig:hardware_multiphase_execution}
\end{figure}

The sledge configuration differed in geometry from the nominal box model used during simulation training, enabling evaluation of policy generalization to previously unseen object shape. All three operational phases were executed sequentially for each payload configuration. The sequential hardware execution of the lifting, transportation, and placement stages under phase-specific policies is illustrated in Fig.~\ref{fig:hardware_multiphase_execution}.

Ground-truth pose of the payload and robot bodies was measured using an OptiTrack motion capture system. Reflective markers attached to the payload and robot modules provided real-time pose estimation. Wheel velocities and joint states were recorded from onboard encoders. For each payload configuration, multiple trials were conducted and all state data were logged for post processing and comparison with simulation results.

At deployment, each unit runs its active phase policy onboard at a fixed control rate from local proprioceptive observations $o_t^{(i)}$, with the relative cargo and neighbor states obtained from joint encoders and the forward kinematics of the rigid grasp. The discrete phase-MDP gate selects the active policy, and the synchronization clamp of Section~\ref{sec:sim2real} is applied online before commands are issued to the wheel and arm actuators. The OptiTrack measurements serve only for offline ground-truth evaluation and the reported metrics; they are not used in the control loop.

\section{Results and Discussion}

\subsection{Policy Performance in Simulation}

Simulation experiments were conducted to evaluate stage-specific performance of the lifting, transportation, and placement policies. Performance was assessed using state-based metrics rather than cumulative reward.
\subsubsection*{Lifting}

\begin{figure}[t]
    \centering
    \begin{minipage}{0.49\columnwidth}
        \centering
        \includegraphics[width=\linewidth]{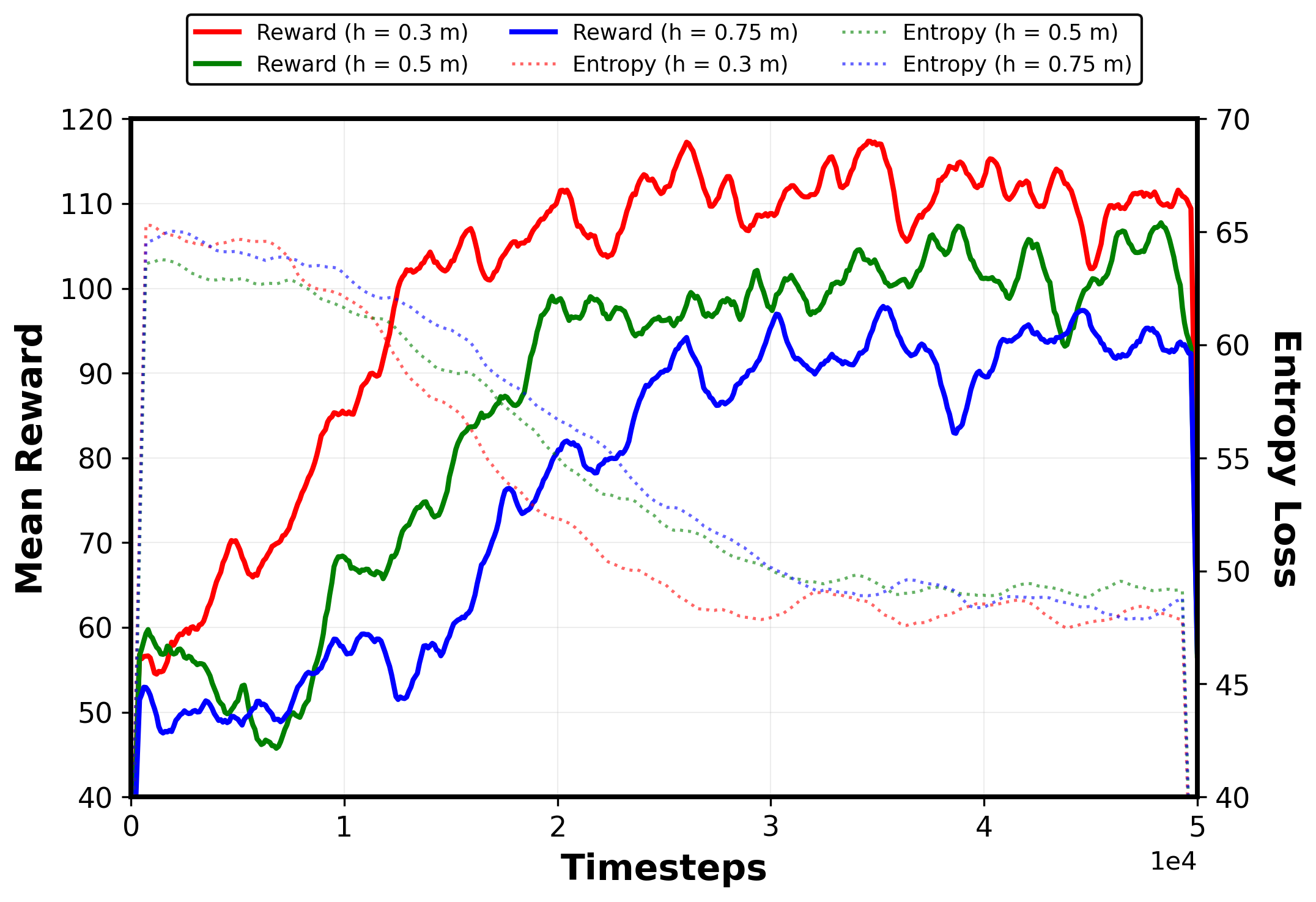}
        {\footnotesize (a)}
    \end{minipage}
    \hfill
    \begin{minipage}{0.49\columnwidth}
        \centering
        \includegraphics[width=\linewidth]{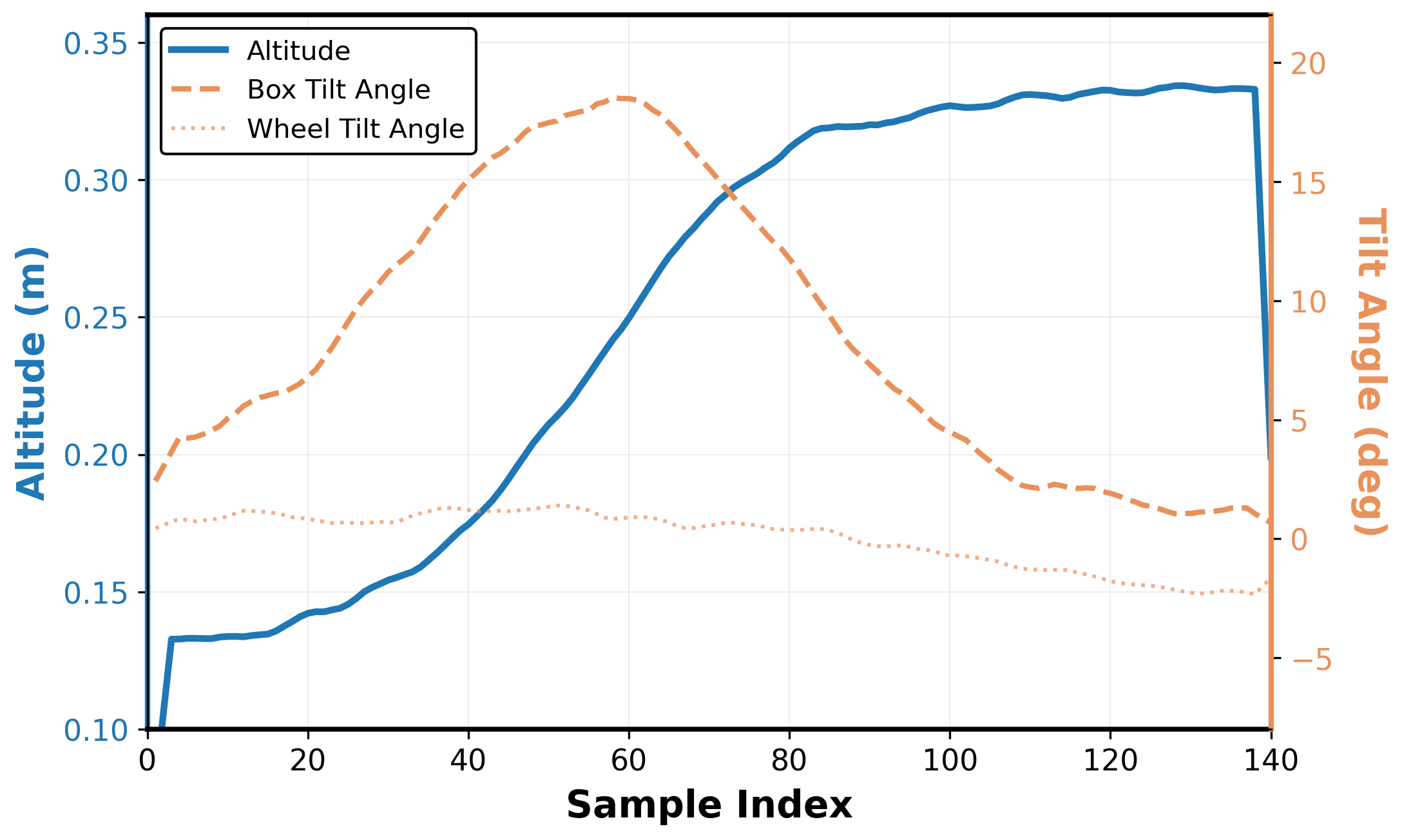}
        {\footnotesize (b)}
    \end{minipage}
    \caption{Simulation results for the lifting policy. (a) PPO training convergence for different target heights. (b) Representative rollout showing payload altitude and tilt evolution during lifting.}
    \label{fig:sim_lifting_results}
\end{figure}

Figure~\ref{fig:sim_lifting_results} reports the learning behavior and execution performance of the lifting policy across target heights of 0.3, 0.5, and 0.75~m.

As shown in Fig.~\ref{fig:sim_lifting_results}(a), reward converges in $1.72\times10^4$, $2.28\times10^4$, and $2.43\times10^4$ timesteps for the 0.3, 0.5, and 0.75~m targets, respectively. The final plateau reward decreases from 105.9 to 94.7 as the lift height increases, reflecting the increased torque demand and sensitivity to coupling at higher elevations. 

Trajectory-level evaluation in Fig.~\ref{fig:sim_lifting_results}(b) shows monotonic height increase without vertical overshoot. The final height tracking error scales with target elevation, remaining at 5.9~mm, 17.4~mm, and 36.0~mm for the 0.3, 0.5, and 0.75~m lifts, respectively. Peak transient tilt occurs during mid-lift torque amplification, where $\theta_{tol}$ acts as a reward-shaping reference rather than a strict safety bound, reaching 17.9$^\circ$, 22.8$^\circ$, and 25.4$^\circ$ across the three targets. Following this transient phase, the payload stabilizes with steady-state tilt bounded below 2.5$^\circ$ in all cases.

\subsubsection*{Transport}

\begin{figure}[t]
    \centering
    \begin{minipage}{0.49\columnwidth}
        \centering
        \includegraphics[width=\linewidth]{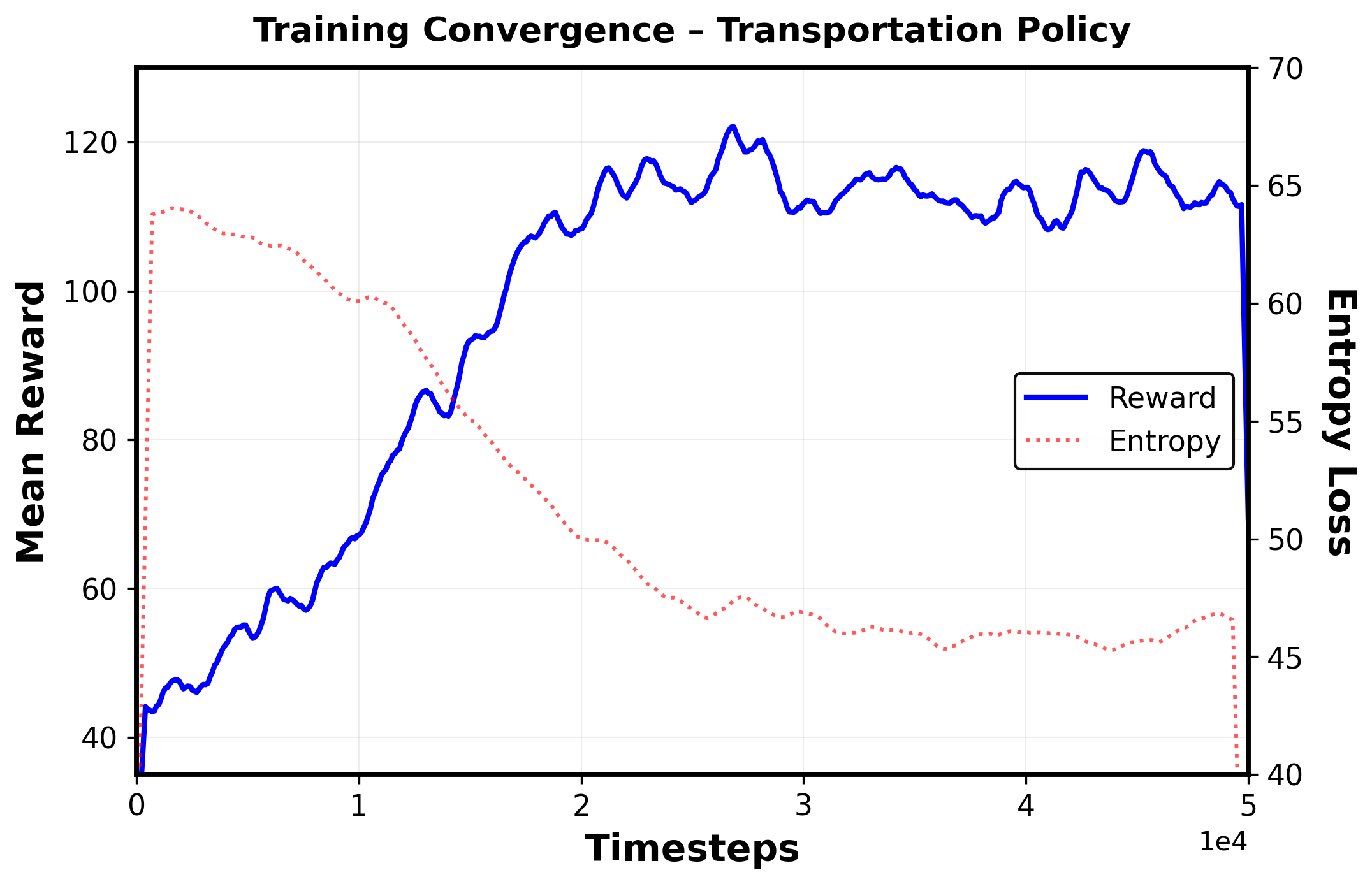}
        {\footnotesize (a)}
    \end{minipage}
    \hfill
    \begin{minipage}{0.49\columnwidth}
        \centering
        \includegraphics[width=\linewidth]{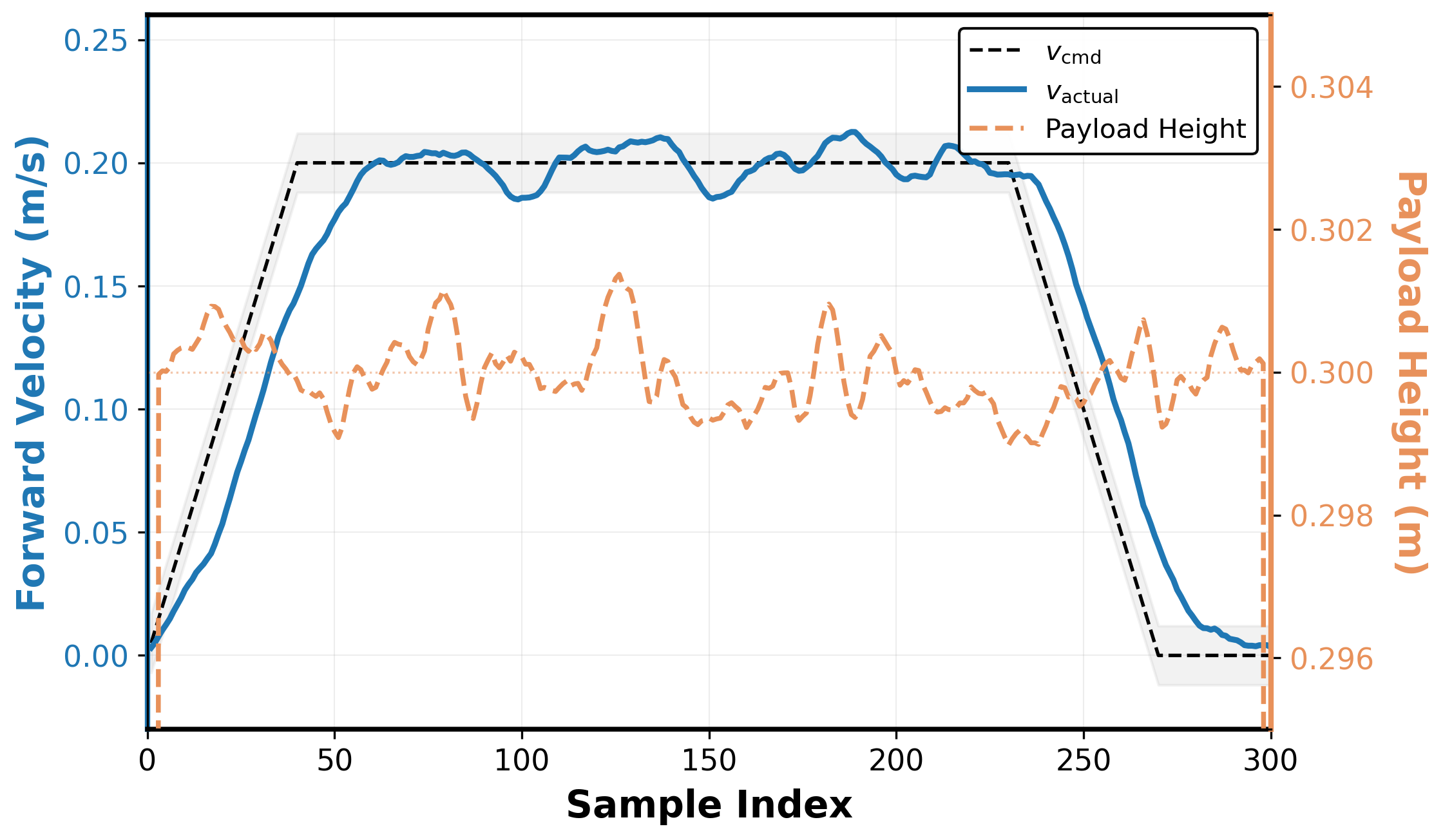}
        {\footnotesize (b)}
    \end{minipage}
    \caption{Simulation results for the transport policy. (a) PPO training convergence showing reward saturation and gradual entropy decay. (b) Forward velocity tracking ($v_{\text{cmd}}$ vs.\ $v_{\text{actual}}$) and concurrent payload height regulation during representative execution.}
    \label{fig:sim_transport_results}
\end{figure}

Figure~\ref{fig:sim_transport_results} summarizes the learning behavior and closed-loop performance of the transport policy.

As shown in Fig.~\ref{fig:sim_transport_results}(a), the mean episodic reward increases steadily and reaches 95\% of its final plateau at approximately $1.74\times10^4$ timesteps, indicating rapid convergence relative to the $5\times10^4$ training horizon. The peak reward overshoot is limited to 8.9\% (122.1 vs.\ 112.1), and the entropy term decays smoothly without collapse.

Figure~\ref{fig:sim_transport_results}(b) evaluates trajectory-level behavior under a 0.2~m/s velocity command. The controller achieves an RMS velocity tracking error of 0.023~m/s (approximately 11\% of command magnitude), with smooth acceleration and deceleration profiles and no oscillatory instability. Simultaneously, the payload height is maintained at 0.30~m with a maximum deviation of 1.58~mm, demonstrating effective decoupling between longitudinal motion and vertical support. 

\subsubsection*{Lowering and Placement}

\begin{figure}[t]
    \centering
    \begin{minipage}{0.49\columnwidth}
        \centering
        \includegraphics[width=\linewidth]{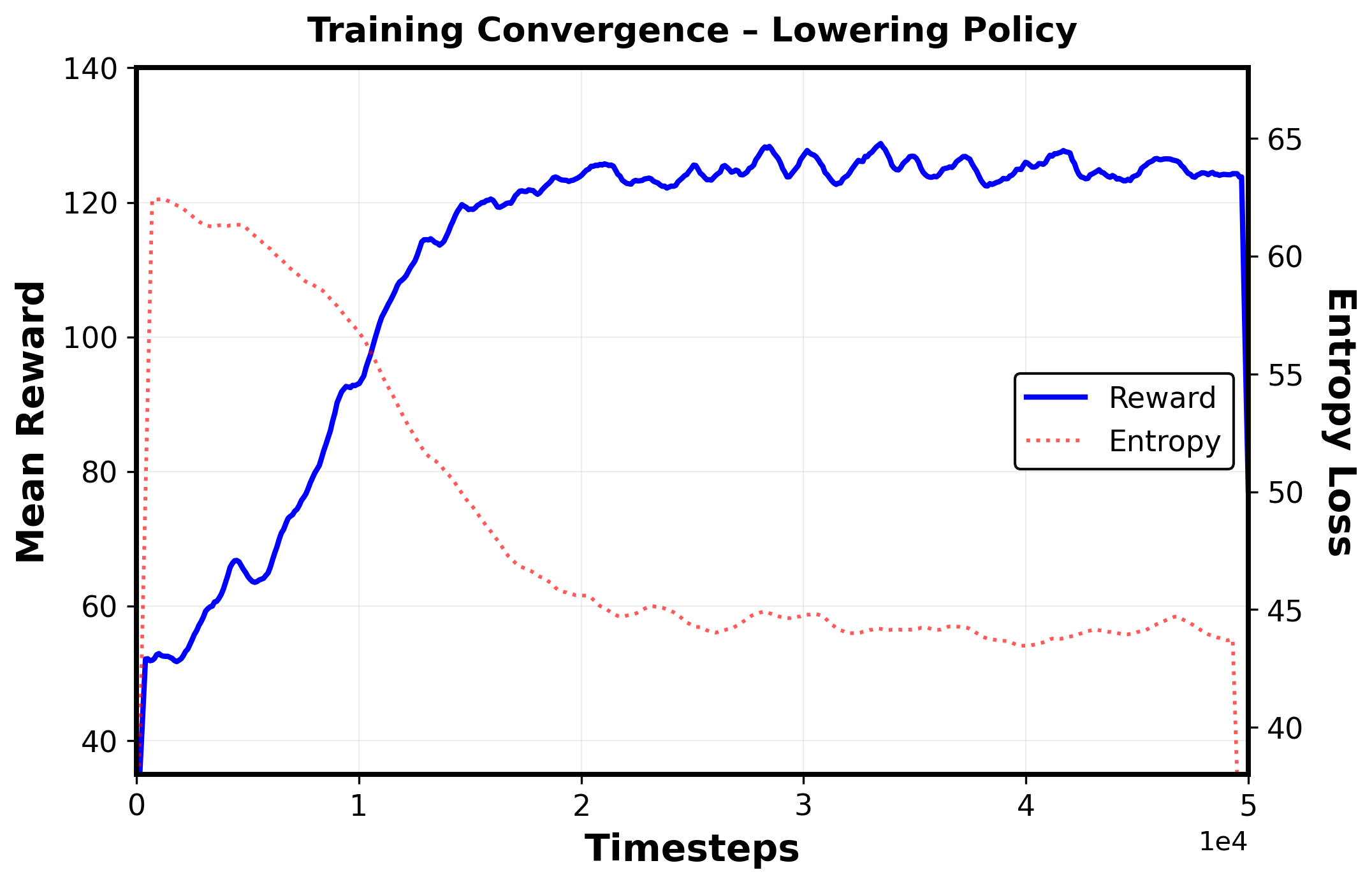}
        {\footnotesize (a)}
    \end{minipage}
    \hfill
    \begin{minipage}{0.49\columnwidth}
        \centering
        \includegraphics[width=\linewidth]{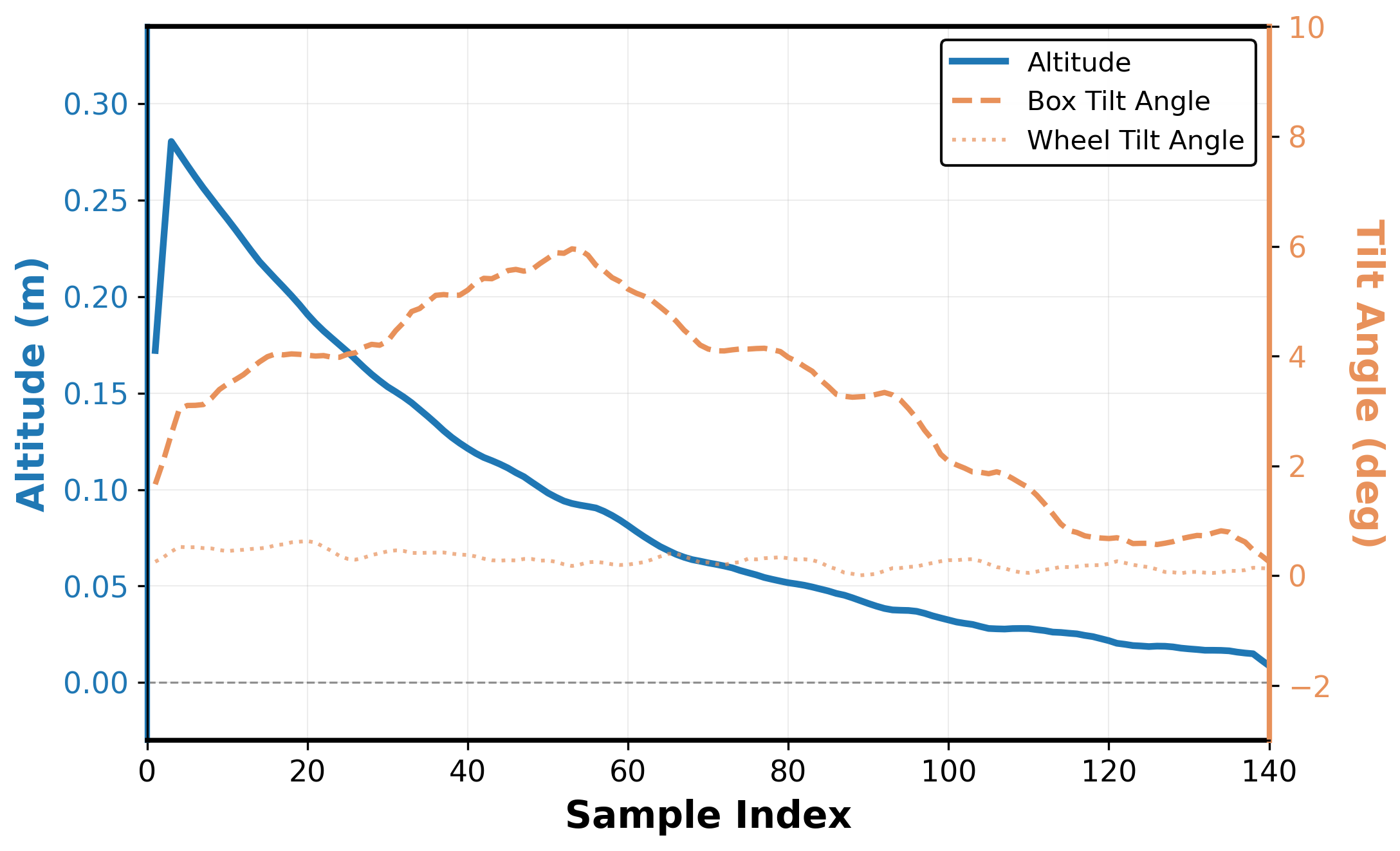}
        {\footnotesize (b)}
    \end{minipage}
    \caption{Simulation results for the lowering policy. (a) PPO training convergence showing rapid reward stabilization and gradual entropy decay. (b) Representative rollout illustrating altitude descent and tilt evolution during payload placement.}
    \label{fig:sim_lowering_results}
\end{figure}

Figure~\ref{fig:sim_lowering_results} presents the learning behavior and execution performance of the lowering policy. As shown in Fig.~\ref{fig:sim_lowering_results}(a), the policy converges within $1.42\times10^4$ timesteps, reaching 95\% of its final plateau faster than the lifting and transport phases. Reward evolution is smooth and entropy decays without instability.

Trajectory-level evaluation in Fig.~\ref{fig:sim_lowering_results}(b) shows controlled descent toward ground contact. The final placement height error is 13.3~mm relative to the ground reference. The impact velocity at contact is 0.030~m/s, which is substantially lower than free-fall velocity from 0.3~m, confirming a soft touchdown rather than ballistic descent. Peak transient tilt during mid-descent is limited to 6.0$^\circ$, and the steady-state tilt after contact is 0.6$^\circ$, indicating stable upright placement. 

\subsection{Baseline Training Without Phase Decomposition}\label{subsec:baseline_no_decomp}

\begin{figure}[t]
\centering

\begin{minipage}{0.49\columnwidth}
\centering
\includegraphics[width=\linewidth]{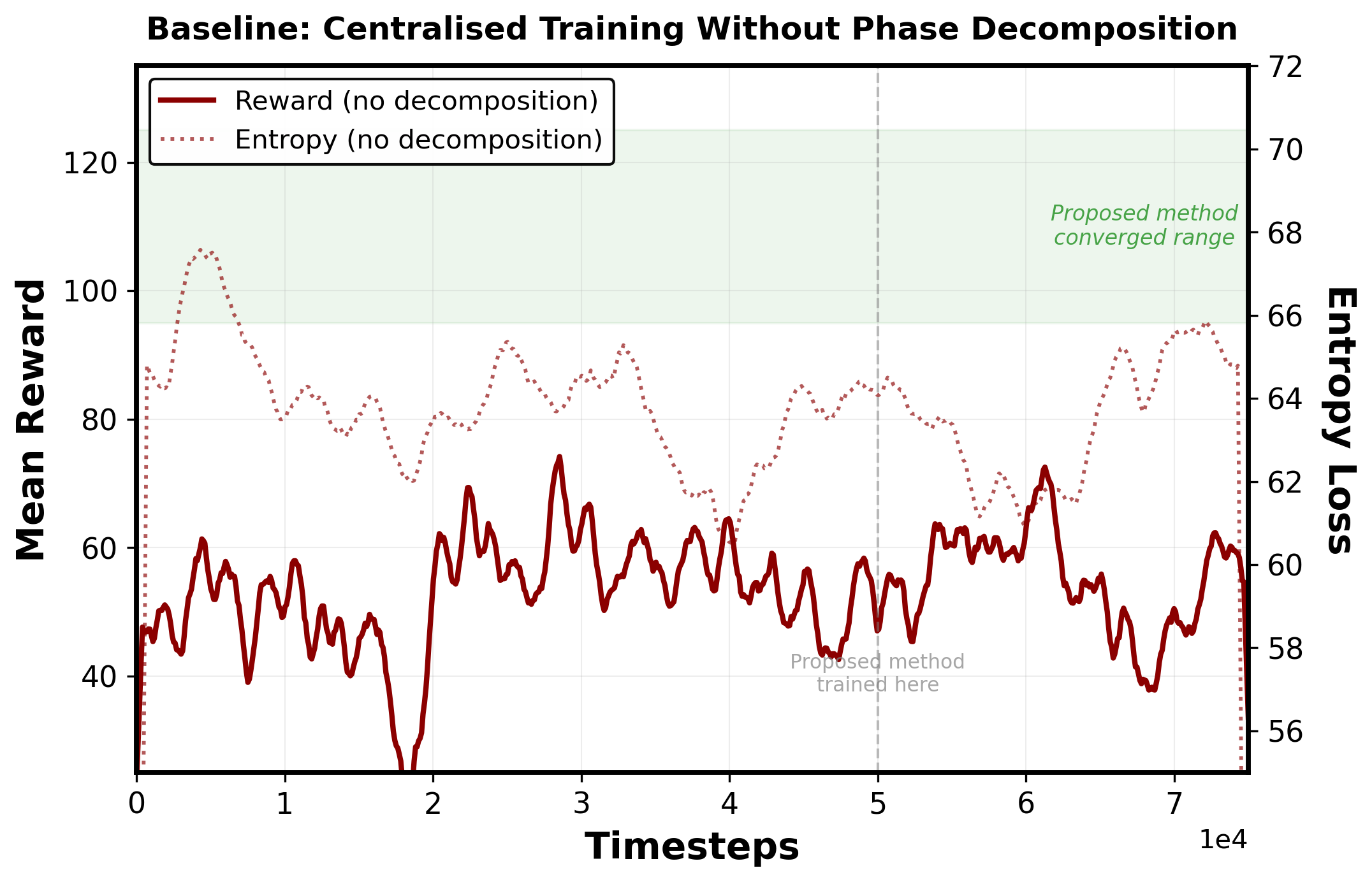}
\end{minipage}
\hfill
\begin{minipage}{0.49\columnwidth}
\centering
\includegraphics[width=\linewidth]{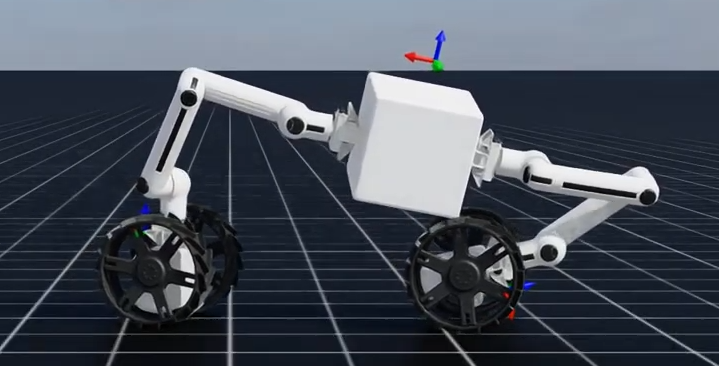}
\end{minipage}

\caption{
Centralized training without phase decomposition.
Left: reward and entropy evolution for a monolithic policy trained over the entire task.
Right: representative simulation snapshot showing unstable behaviour where translation occurs during lifting or placement.
}
\label{fig:baseline_failure}

\end{figure}

To evaluate the necessity of phase decomposition and joint sync, we trained a single centralized policy for the full mission without stage separation or synchronization. As shown in Fig.~\ref{fig:baseline_failure}, the monolithic policy fails to converge even after $7.5\times10^{4}$ timesteps, which exceeds the training horizon used for the individual phase policies. The reward remains oscillatory and entropy does not exhibit sustained decay, indicating that the controller fails to learn a stable strategy. Qualitative simulation behaviour further shows unstable motion where the robot attempts to move while lifting or placing the payload simultaneously. These results suggest that phase decomposition is necessary for stable learning under configuration-dependent coupling.

\subsection{Evaluation on Hardware}

\begin{figure}[!t]
    \centering
    \includegraphics[width=\linewidth]{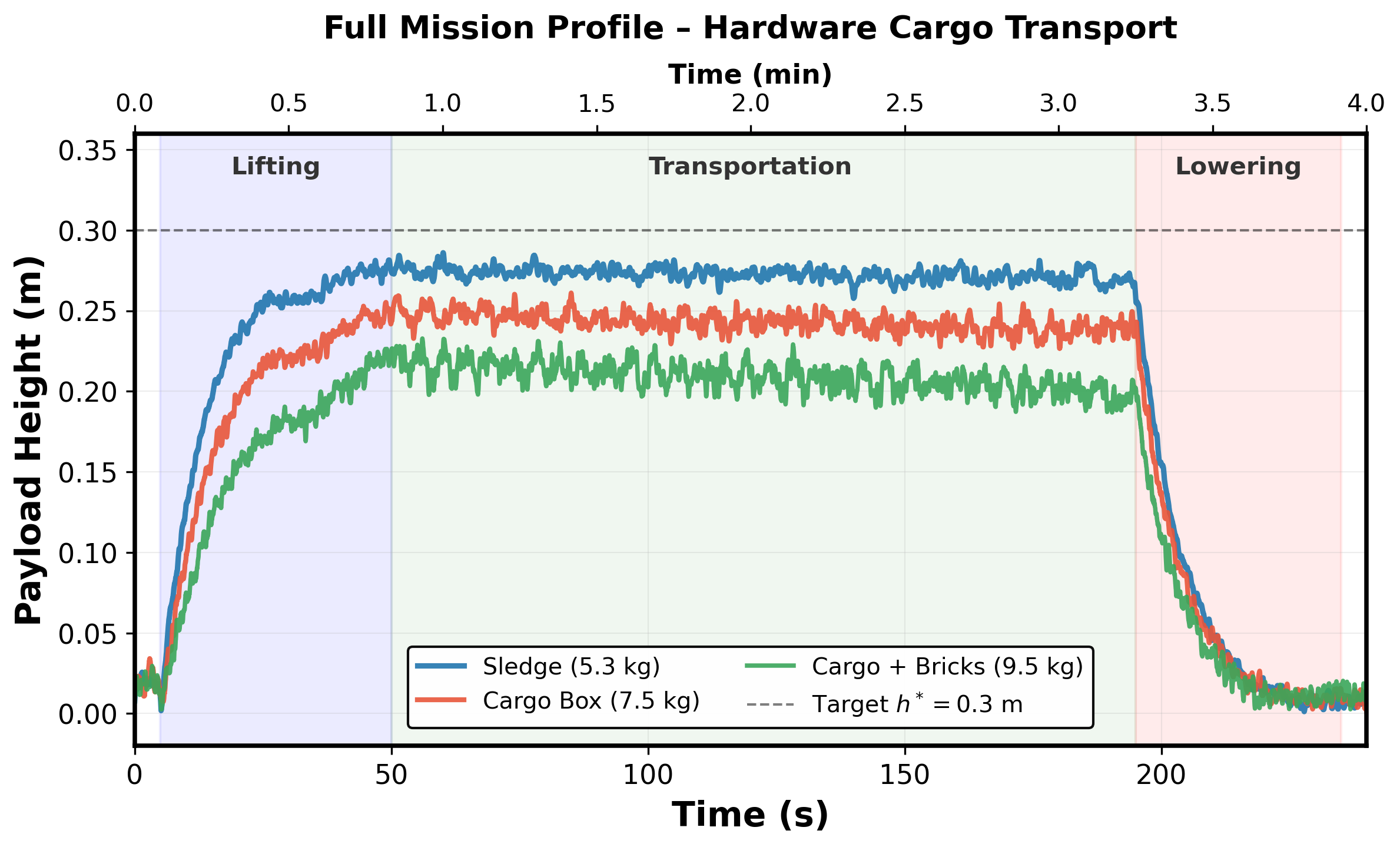}
    \caption{Full hardware mission profile showing lifting,
    transportation, and lowering for three payload masses.
    Curves represent mean height over three trials per payload.
    The dashed line indicates the 0.30 m reference height.}
    \label{fig:hw_full_mission}
\end{figure}

Hardware validation was conducted for three payload configurations (5.3 kg, 7.5 kg, and 9.5 kg), with three trials per payload. Figure~\ref{fig:hw_full_mission} shows the averaged payload height profile across the full mission (lifting, transport, and lowering). All values reported below are mean $\pm$ 1$\sigma$ over trials.

During transportation, the steady-state payload height decreases monotonically with mass, measuring
$272.6 \pm 3.8$ mm (5.3 kg),
$243.0 \pm 5.5$ mm (7.5 kg), and
$209.1 \pm 7.8$ mm (9.5 kg).
Relative to the 0.30 m target, these correspond to
approximately 91\%, 81\%, and 70\% height retention,
respectively.
The increased undershoot and variance with payload mass
are consistent with terrain sinkage and amplified arm torque
coupling under heavier loading.

The time required to reach 95\% of steady-state height also
increases with mass, from
$21.02 \pm 0.61$ s (5.3 kg) to
$30.42 \pm 0.76$ s (7.5 kg) and
$31.18 \pm 0.56$ s (9.5 kg),
indicating slower vertical convergence under higher load.

During placement, all payloads exhibit controlled descent
with small residual post-contact height.
The final settled height after lowering is
$8.1 \pm 0.7$ mm,
$8.0 \pm 1.0$ mm, and
$12.1 \pm 1.0$ mm
for increasing mass, respectively.
The slightly higher residual height for the heaviest payload
is attributable to deeper sand penetration and local terrain
deformation.

Hardware results confirm stable multi-phase execution across payload masses. While absolute lifting height reduces under heavier payloads, vertical regulation remains consistent and bounded across trials, demonstrating robust completion with degraded height regulation under heavier loads in granular terrain conditions.

\section{CONCLUSIONS}
This work presented a phase-decomposed reinforcement learning framework for cooperative lunar cargo transport using physically reconfigurable wheel-arm robotic units. Stage-specific policies were trained in simulation and validated through lunar-analog hardware experiments, demonstrating stable multi-phase execution under varying payload masses and terrain conditions. The results indicate that phase decomposition with synchronized execution improves robustness to configuration-dependent dynamics and sim-to-real discrepancies.

\bibliographystyle{./IEEEtran}
\bibliography{reference}

\end{document}